\pdfoutput=1
\documentclass[11pt]{article}

\usepackage[final]{acl}

\usepackage{times}
\usepackage{latexsym}

\usepackage[T1]{fontenc}
\usepackage[utf8]{inputenc}

\usepackage{microtype}

\usepackage{inconsolata}

\usepackage{graphicx}

\usepackage{dsfont}
\usepackage{bm}
\usepackage{bbm}
\usepackage{graphicx}
\usepackage{color}
\usepackage{multicol}
\usepackage{multirow}
\usepackage{wrapfig,lipsum,booktabs}
\usepackage[ruled,vlined,linesnumbered]{algorithm2e}
\usepackage{pgfplots}
\pgfplotsset{compat=1.12} 
\usepackage{filecontents}
\usepackage{tikz}
\usepackage{xcolor}
\usepackage{colortbl}
\usepackage{xspace}
\usepackage{makecell}
\usepackage{soul}
\usepackage{amsmath,amsfonts,amssymb}
\usepackage{subcaption}
\usetikzlibrary{calc}
\usepgfplotslibrary{groupplots}
\usetikzlibrary{angles,quotes} 
\usetikzlibrary{shapes,arrows}
\usetikzlibrary{backgrounds}
\usetikzlibrary{matrix}
\usetikzlibrary{patterns}
\usepackage{tikz-3dplot}
\usepackage{hyperref}
\usepackage{cleveref}
\usepackage{paralist}
\usepackage{cancel}
\usepackage{todonotes}
\usepackage{tabu}
\usepackage{rotating}
\usepackage{etoolbox}
\usepackage{adjustbox}
\usepackage{enumerate}
\usepackage{enumitem}
\setitemize{itemsep=0pt,topsep=0pt,parsep=0pt,partopsep=0pt}
\setenumerate{itemsep=0pt,topsep=0pt,parsep=0pt,partopsep=0pt}
\usepackage{pifont}
\usepackage{cancel}
\usepackage{lipsum}
\usepackage{listings,lstautogobble}
\usepackage{fancyvrb}
\usepackage{fvextra}
\usepackage{caption}
\usepackage{arydshln}

\newcommand{\model}[1]{\textsc{#1}\xspace}

\newcommand{\qwen}{\model{Qwen1.5-0.5B}}
\newcommand{\gemma}{\model{Gemma-2B}}
\newcommand{\llama}{\model{Llama-3-8B}}
\newcommand{\prop}{\model{Prop.}}
\newcommand{\temp}{\model{Temp.}}
\newcommand{\uni}{\model{Uni.}}
\newcommand{\cossim}{\model{CosSim}}
\newcommand{\diff}{\model{Diff}}
\newcommand{\ema}{\model{EMA}}

\newcommand{\ours}{\model{MoS}}
\newcommand{\oursfull}{\model{Mixture-of-Skills}}
\newcommand{\ourspec}{\model{MoSpec}}
\newcommand{\mathllama}{\model{MathLlama-3-8B}}

\newcommand{\multidds}{\model{MultiDDS}}
\newcommand{\multiuat}{\model{MultiUAT}}

\newcommand{\dataset}[1]{\texttt{#1}\xspace}
\newcommand{\mmlu}{\dataset{MMLU}}
\newcommand{\mtbench}{\dataset{MT-bench}}

\newcommand{\avgmmlu}{\mu_{\textrm{MU}}}
\newcommand{\avgmtbench}{\mu_{\textrm{MB}}}
\newcommand{\avgboth}{\mu_{\textrm{BOTH}}}
\newcommand{\avgall}{\mu_{\textrm{ALL}}}

\newcommand{\dtrain}{D_{\textrm{trn}}}
\newcommand{\dvalid}{D_{\textrm{dev}}}

\newcommand{\vscorer}{\pmb{\psi}}
\newcommand{\vx}{\pmb{x}}
\newcommand{\vy}{\pmb{y}}
\newcommand{\vh}{\pmb{h}}
\newcommand{\ve}{\pmb{e}}
\newcommand{\vz}{\pmb{z}}

\newcommand{\vtheta}{\pmb{\theta}}

\newcommand*{\affmark}[1][*]{\textsuperscript{#1}}

\DeclareMathOperator*{\argmin}{argmin}

\title{\oursfull: Learning to Optimize Data Usage for Fine-Tuning Large Language Models}

\author{
  Minghao Wu\affmark[$\heartsuit$]\enskip  Thuy-Trang Vu\affmark[$\heartsuit$]\enskip Lizhen Qu\affmark[$\heartsuit$]\enskip Gholamreza Haffari\affmark[$\heartsuit$] \\
  \affmark[$\heartsuit$]Monash University \\
  \texttt{\{firstname.lastname\}@monash.edu}
}

\begin{document}

\renewcommand{\tableautorefname}{Table}
\renewcommand{\sectionautorefname}{Section}
\renewcommand{\subsectionautorefname}{Section}
\renewcommand{\subsubsectionautorefname}{Section}
\renewcommand{\figureautorefname}{Figure}
\renewcommand{\equationautorefname}{Equation}
\renewcommand{\algorithmautorefname}{Algorithm}
\newcommand{\linenoautorefname}{Line}

\maketitle
\begin{abstract}

Large language models (LLMs) are typically fine-tuned on diverse and extensive datasets sourced from various origins to develop a comprehensive range of skills, such as writing, reasoning, chatting, coding, and more. Each skill has unique characteristics, and these datasets are often heterogeneous and imbalanced, making the fine-tuning process highly challenging. Balancing the development of each skill while ensuring the model maintains its overall performance requires sophisticated techniques and careful dataset curation. In this work, we propose a general, model-agnostic, reinforcement learning framework, \oursfull (\ours), that learns to optimize data usage automatically during the fine-tuning process. This framework ensures the optimal comprehensive skill development of LLMs by dynamically adjusting the focus on different datasets based on their current learning state. To validate the effectiveness of \ours, we conduct extensive experiments using three diverse LLM backbones on two widely used benchmarks and demonstrate that \ours substantially enhances model performance. Building on the success of \ours, we propose \ourspec, an adaptation for task-specific fine-tuning, which harnesses the utilities of various datasets for a specific purpose. Our work underlines the significance of dataset rebalancing and present \ours as a powerful, general solution for optimizing data usage in the fine-tuning of LLMs for various purposes.

\end{abstract}

\section{Introduction}

Large language models (LLMs) have demonstrated their extraordinary capabilities and are expected to proficiently master a diverse range of skills \citep{DBLP:conf/nips/Ouyang0JAWMZASR22, DBLP:conf/iclr/SanhWRBSACSRDBX22,DBLP:journals/corr/abs-2303-08774, DBLP:journals/corr/abs-2305-10403, DBLP:journals/corr/abs-2302-13971,DBLP:journals/corr/abs-2307-09288,DBLP:journals/corr/abs-2312-11805,DBLP:journals/corr/abs-2403-08295}, such as writing, reasoning, chatting, coding, and more, through supervised fine-tuning (SFT) and reinforcement learning from human feedback (RLHF) on an extensive collection of datasets from various sources \citep{DBLP:journals/corr/abs-2204-05862,DBLP:conf/icml/LongpreHVWCTZLZ23,ding-etal-2023-enhancing,wang-etal-2024-assessing}. Each dataset contributes unique elements to the model's skill set, but this diversity also brings challenges.

\begin{figure}[t]
\centering

\begin{tikzpicture}
    % Nodes
    \node[fill=red!20, rectangle, rounded corners=3pt] (dtrain) at (0, 1) {\small Training Data $\dtrain$};
    \node[fill=blue!20, rectangle, rounded corners=3pt] (scorer) at (0, 0) {\small Scorer $\vscorer$};
    \node[fill=blue!20, rectangle, rounded corners=3pt] (model) at (5, -2) {\small LLM $\vtheta$};
    
    \node[] (dataset) at (2.5,0) {\small $\tilde{i} \sim p_{\vscorer}(N)$};
    \node[] (batch) at (5,-1) {\small $(\vx, \vy) \sim \dtrain^{\tilde{i}}$};

    \node[] (rbatch) at (2.5,-2) {\small $(\vx, \vy) \sim \dtrain^{i}$};
    \node[] (reward) at (0,-1) {\small Reward $\mathcal{R}_{\{\cdot\}}(i)$};

    % Arrows
    \draw[->, line width=1.5pt] (dtrain) -- (scorer);
    \draw[->, line width=1.5pt] (scorer.east) -- (dataset.west);
    \draw[->, line width=1.5pt] (dataset.east) to [in=90, out=0] (batch.north);
    \draw[->, line width=1.5pt] (batch.south) -- (model.north);

    \draw[->, dashed, line width=1.5pt] (model.west) -- (rbatch.east);
    \draw[->, dashed, line width=1.5pt] (rbatch.west) to [out=180, in=270] (reward.south);
    \draw[->, dashed, line width=1.5pt] (reward.north) -- (scorer.south);

\end{tikzpicture}

\caption{
The overview of \oursfull. 
The training collection $\dtrain = \{\dtrain^i\}_{i=1}^{N}$ consists of various SFT datasets, with $\dtrain^{i}$ indicating the $i$-th dataset.
Please refer to \autoref{sec:method} for more details.
}
\label{fig:overview}
\end{figure}
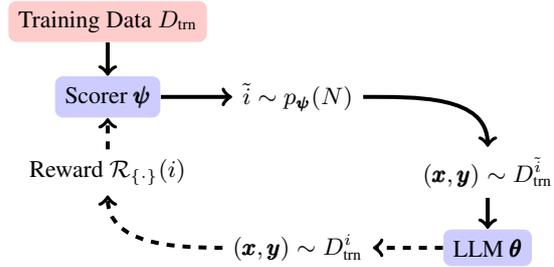

One common challenge in fine-tuning models across multiple datasets is dealing with their \textbf{heterogeneity} (where different datasets exhibit distinct characteristics) and \textbf{imbalance} (where datasets vary in size and accessibility), making the fine-tuning process highly challenging. To address this challenge, recent approaches often cap dataset usage to prevent models from being overwhelmed by large datasets, but this limits the utilization of all available data \citep{DBLP:journals/jmlr/RaffelSRLNMZLL20,DBLP:journals/corr/abs-2210-11416,DBLP:journals/corr/abs-2212-12017,DBLP:conf/iclr/WeiBZGYLDDL22}. Previous multilingual research adjusts dataset distribution heuristically with a temperature term $\tau$, which requires extensive hyperparameter tuning and overlooks dataset interactions and model learning dynamics \citep{DBLP:journals/corr/abs-1907-05019,conneau-etal-2020-unsupervised}. This leads us to a critical research question: \textit{Is there a better way to optimize the data usage?}

Building on the research question posed, we first confirm that adjusting the dataset usage properly can significantly enhance model performance (see \autoref{tab:main}). Moreover, inspired by Differentiable Data Selection \citep{DBLP:conf/icml/WangPMACN20}, we propose a general, model-agnostic reinforcement learning framework \oursfull (\ours) that \textit{learns} to optimize data usage automatically during the fine-tuning process. To achieve this optimization, we introduce another set of parameters, $\vscorer$, known as the scorer network. As shown in \autoref{fig:overview}, this network dynamically adjusts data usage based on the current learning state of the LLM $\vtheta$. Furthermore, the rewards used to update the scorer network $\vscorer$ are provided by the LLM $\vtheta$ from three distinct perspectives: \textit{transferability}, \textit{difficulty}, and \textit{learning trajectory}, ensuring that the scorer network can effectively guide the data usage optimization process. All these efforts constitute the success of our \oursfull framework.

To validate the effectiveness of \ours, we conduct extensive experiments using three diverse model backbones: \qwen \citep{DBLP:journals/corr/abs-2309-16609}, \gemma \citep{DBLP:journals/corr/abs-2403-08295}, and \llama \citep{llama3modelcard}, on two widely-used benchmarks: \mmlu \citep{DBLP:conf/iclr/HendrycksBBZMSS21} and \mtbench \citep{DBLP:journals/corr/abs-2306-05685}. Our empirical results demonstrate that \ours substantially improves the models' overall performance. Our analysis indicates that our model not only effectively learns optimal data utilization but also accelerates training convergence by $2.2 \times$. Additionally, it demonstrates robustness against variations in sampling priors and integrates seamlessly with advanced instance selection methods.
% These benefits contribute to the overall performance improvements observed in our experiments. 
Furthermore, we explore the application of \ours in task-specific fine-tuning. We show that \ours, with minor reward modifications known as \ourspec, can be effectively used for task-specific fine-tuning. 

Our contributions are summarized as follows:
\begin{itemize}
    \item We present a general, model-agnostic reinforcement learning framework, \oursfull (\ours), that learns to automatically optimize data usage during the SFT process with three novel rewards (see \autoref{sec:method}).
    \item Extensive experiments with three model backbones on two benchmarks demonstrate that \ours significantly enhances model performance. Our analysis reveals that \ours not only effectively learns the optimal data usage but also accelerates training convergence by $2.2\times$. Additionally, it maintains robustness against variations in sampling priors and is compatible with strong instance selection methods (see \autoref{sec:experiments} and \autoref{sec:analysis}).
    \item We explore the application of \ours in task-specific fine-tuning, introducing a variant called \ourspec. This variant, with minor modifications to the rewards, is proven to effectively harness diverse datasets for task-specific fine-tuning (see \autoref{sec:ourspec}).
\end{itemize}

\section{Preliminaries}
\label{sec:preliminaries}

\paragraph{Supervised Fine-Tuning}
A large language model (LLM), parameterized by $\vtheta$, is capable of following and responding to human instructions after supervised fine-tuning (SFT). Given a single training dataset $\dtrain^{1} = \{ (\vx_{j}, \vy_{j}) \}_{j=1}^{M_{1}}$, where $M_{1}$ is the size of $\dtrain^{1}$ and $\vx_j$ and $\vy_j$ are the instruction and response of the $j$-th example, the objective function during SFT is to minimize the negative log-likelihood with respect to $\vtheta$:
\begin{align}
    \mathcal{L}_{s}(\dtrain^{1}; \vtheta) = - \sum^{M_1}_{j=1} \log p(\vy_j | \vx_j; \vtheta).
\end{align}
When fine-tuning the LLM $\vtheta$ over multiple datasets $\dtrain = \{\dtrain^i\}_{i=1}^{N}$, where $\dtrain^{i} = \{ (\vx^{i}_{j}, \vy^{i}_{j}) \}_{j=1}^{M_i}$, the objective function becomes:
\begin{align}
\label{eq:naive_multi_loss}
    \mathcal{L}(\dtrain; \vtheta) = \sum^{N}_{i=1} \mathcal{L}_{s}(\dtrain^{i}; \vtheta).
\end{align}

\paragraph{Heuristic Balancing by Temperature}
Instead of merging all datasets into a single training mixture, a common approach is to adjust the sampling probability of texts in different languages using a temperature term $\tau$ \citep{DBLP:journals/corr/abs-1907-05019,conneau-etal-2020-unsupervised}. Specifically, the sampling probability of the $i$-th dataset is $q(i) = \frac{|M_i|}{\sum^{N}_{n=1}|M_n|}$ and can be adjusted by the temperature $\tau$ as:
\begin{align}
    q_\tau(i) = \frac{q(i)^{1/\tau}}{\sum_{n=1}^{N} q(n)^{1/\tau}}.
    \label{eq:temp}
\end{align}
Consequently, $\tau = 1$ corresponds to proportional sampling, equivalent to \autoref{eq:naive_multi_loss}. Conversely, $\tau = \infty$ corresponds to uniform sampling, where smaller datasets are up-sampled to match the largest dataset. The loss function becomes:
\begin{align}
    \mathcal{L}(\dtrain; \vtheta, q_\tau(N)) = \mathbb{E}_{i \sim q_\tau(N)} \left[\mathcal{L}_{s}(\dtrain^{i}; \vtheta)\right].
\end{align}

\paragraph{Differentiable Data Selection (\model{DDS})}
\citet{DBLP:conf/icml/WangPMACN20} propose a general framework that automatically re-weighs training instances to enhance model performance, utilizing a validation set $\dvalid$. This framework consists of two components: the model $\vtheta$ and the scorer network $\vscorer$. The scorer network $\vscorer$ is designed to calculate a sampling probability for each training instance, which reflects its impact on validation performance. Training instances that have a greater similarity with $\dvalid$ are allocated a higher probability, thus increasing their likelihood of being selected for model $\vtheta$ updates. \ours is inspired by \model{DDS} but has key differences. Firstly, \ours focuses on rebalancing datasets, unlike \model{DDS}, which reweighs training instances. Secondly, \ours does not require prior knowledge of downstream applications, whereas DDS relies on validation set feedback, risking overfitting to that specific validation set. Thirdly, \model{DDS} uses the same architecture for the scorer network $\vscorer$ and model $\vtheta$, limiting its scalability, while \ours uses a simple MLP model as its scorer network (see \autoref{sec:mos}).

\section{\oursfull}
\label{sec:method}

In this section, we provide a detailed overview of \oursfull (\ours). We begin by outlining the reinforcement learning framework employed in \ours (\autoref{sec:mos}). Following this, we discuss the reward functions (\autoref{sec:rewards}).

\begin{algorithm}[t]
    \SetKwInOut{Input}{Input}
    \SetKwInOut{Output}{Output}
    \Input{$\dtrain =\{ \{ (\vx_{j}^{i}, \vy_{j}^{i}) \}^{M_i}_{j=1} \}^{N}_{i=1}$, $N$ training datasets with the size of $M_i$ for the $i$-th dataset; $S$, update frequency of $\vscorer$; $T$, total training steps; $\alpha$, learning rate for $\vtheta$; $\gamma$, learning rate for $\vscorer$;}
    \Output{The converged model $\vtheta$;}
    \SetAlgoLined
    Initialize $p_{\vscorer_{0}}(N)$ as \autoref{eq:temp} with $\tau=\infty$\; \label{line:line1}
    \For{t=0 to T}{
        $\tilde{i} \sim p_{\vscorer}(N)$\;
        Sample batch $(\vx, \vy) \sim \dtrain^{\tilde{i}}$\;
        $\vtheta \leftarrow \vtheta - \alpha \cdot \nabla_{\vtheta} \mathcal{L}(\vy|\vx ; \vtheta) $\; \label{line:line5}
        \If{t \% $S$ == 0}{ \label{line:line6}
            \For{i=1 to N}{
                $(\vx', \vy') \sim \dtrain^i$\;
                Compute reward $\mathcal{R}_{\{\cdot\}}(i)$ for $\dtrain^i$ as in \autoref{sec:rewards}\; \label{line:line9}
            }
            $\vscorer \leftarrow \vscorer + \sum_{i=1}^{N} \gamma \cdot \mathcal{R}_{\{\cdot\}}(i) \cdot \nabla_{\vscorer} \log p_{\vscorer}(i)$\label{line:line11}
        }
    }
    \caption{\oursfull}
    \label{alg:mos}
\end{algorithm}
\subsection{Learning to Optimize Data Usage}
\label{sec:mos}

We propose \oursfull (\ours) that \textit{learns} to optimize the data usage during the fine-tuning process by training a scorer network, parameterized by $\vscorer$,  within a reinforcement learning (RL) framework. In this setup, the LLM $\vtheta$ and the training datasets $\dtrain$ constitute the \textit{environment}, while our scorer network serves as the RL \textit{agent}. In this framework, unlike the static sampling probabilities described in \autoref{eq:temp}, the scorer network $\vscorer$ dynamically adjusts the sampling probabilities for each dataset in $\dtrain$ according to the current learning state of the LLM $\vtheta$. Alternately, the LLM $\vtheta$ is optimized based on the sampling distribution given by the scorer network $\vscorer$.

To provide a broader perspective, \ours can be conceptualized as the resolution of a \textit{bi-level optimization} problem \citep{DBLP:journals/anor/ColsonMS07}. In this view, the outer level optimizes the parameters of the LLM $\vtheta$, while the inner level focuses on adjusting the sampling probabilities through the scorer network $\vscorer$. Hence, the training objective becomes:
\begin{equation}
    \label{eq:bilevel}
    \begin{aligned}
        \vscorer &= \argmin_{\vscorer} {\mathcal{J}(\dtrain; \vtheta(\vscorer))}, \text{where}  \\
        \vtheta(\vscorer) &= \argmin_{\vtheta}\mathbb{E}_{i \sim p_{\vscorer}(N)} [\mathcal{L}(\dtrain^i;\vtheta)].
    \end{aligned}
\end{equation}

Specifically, we present the algorithm of \ours in \autoref{alg:mos}. \ours initially parameterizes the initial sampling probability distribution with $\vscorer$ as shown in \autoref{eq:temp}, using a warm-up temperature $\tau = \infty$ (see \autoref{line:line1}). When updating the LLM $\vtheta$, we employ the standard gradient-based optimization method (see \autoref{line:line5}). For computational efficiency, the scorer network $\vscorer$ is updated every $S$ steps (see \autoref{line:line6}). During updates of $\vscorer$, we randomly draw one mini-batch from each training set $\{\dtrain^i\}_{i=1}^{N}$ and compute the corresponding rewards as described in \autoref{sec:rewards} (see \autoref{line:line9}). The training dataset $\dtrain^{i}$ that yields a high reward is considered to be relatively more beneficial to the overall performance, and its corresponding sampling probability is increased (see \autoref{line:line11}).

A critical issue in \autoref{alg:mos} is that \autoref{eq:bilevel} is not directly differentiable with respect to $\vscorer$. To address this, reinforcement learning (RL) with suitable reward functions is needed \cite{DBLP:conf/icml/WangPMACN20}. The update rule for $\vscorer$ becomes:
\begin{align}
    \vscorer \leftarrow \vscorer + \sum_{i=1}^{N}\mathcal{R}_{\{\cdot\}}(i) \cdot \nabla_{\vscorer} \log p_{\vscorer}(i).
\end{align}
Details for the rewards $\mathcal{R}_{\{\cdot\}}(i)$ are in \autoref{sec:rewards} and the update of the scorer network $\vscorer$ follows the \model{REINFORCE} algorithm \citep{DBLP:journals/ml/Williams92}.

\paragraph{Implementing the scorer network}
Given that the scorer network $\vscorer$ is primarily designed to model a relatively simple distribution over the training datasets $\dtrain$, we utilize a fully connected 2-layer perceptron network for this task. The network takes as input a vector that specifies which training datasets are accessible. Note that the scorer network $\vscorer$ is used for adjusting the data usage during the fine-tuning process and is orthogonal to the reward model used in RLHF.

\subsection{Rewards for Learning}
\label{sec:rewards}

We design the rewards of \ours from three perspectives: transferability (\autoref{sec:transfer}), difficulty (\autoref{sec:difficult}), and learning trajectory (\autoref{sec:trajectory}).

\subsubsection{Transferability}
\label{sec:transfer}
Transferring knowledge from one problem to another related problem is beneficial, and this transferability is often measured by the similarity between datasets \citep{DBLP:journals/corr/abs-1812-02224,DBLP:journals/pieee/ZhuangQDXZZXH21}. Datasets with higher similarity are more likely to make more contributions to the targeted performance of the model.

In this work, we represent the training datasets $\dtrain$ using the mini-batch embeddings and then calculate the pairwise cosine similarities among the mini-batch embeddings for each dataset. We draw a random mini-batch $B^{i} = \{ (\vx^{i}_{j}, \vy^{i}_{j}) \}_{j=1}^{L}$ from $\dtrain^{i}$, where $L$ is the batch size, and the mini-batch embedding $\vz^{i}$ is defined as:
\begin{align}
    \vz^{i} = \frac{1}{L}\sum^{L}_{j=1} \ve^{i}_{j},\quad  \ve^{i}_{j}=\frac{1}{K}\sum^{K}_{k=1} \vh_k,
\end{align}
where $K$ is the sequence length of the concatenation of $\vx^{i}_{j}$ and $\vy^{i}_{j}$, and $\vh_k$ is the hidden state of the token $w_k$ in the concatenated sequence from the topmost layer of the LLM $\vtheta_t$. Consequently, we define the reward $\mathcal{R}_{\cossim}(i)$ for $\dtrain^{i}$ as the average cosine similarity among all training datasets:
\begin{align}
    \mathcal{R}_{\cossim}(i) = \frac{1}{N}\sum^{N}_{n=1} \frac{\vz^{i}\cdot\vz^{n}}{\left\lVert \vz^{i} \right\rVert \cdot \left\lVert \vz^{n} \right\rVert},
\end{align}
where $N$ is the number of datasets in $\dtrain$.

\subsubsection{Difficulty}
\label{sec:difficult}

Recent work demonstrates that the transfer of knowledge between datasets is not always guaranteed \citep{wu-etal-2021-uncertainty}. In response, we attempt to design the reward based on the inherent difficulty of the dataset in this section. 

Recently, the perplexity is used for measure the dataset difficulty \citep{DBLP:journals/corr/abs-2308-12032,DBLP:journals/corr/abs-2309-04564}. Given a training example $(\vx^{i}_{j}, \vy^{i}_{j})$ from $\dtrain^{i}$ and the LLM $\vtheta$, the perplexity is defined as:
\begin{equation}
    \begin{aligned}
        \text{PPL}&(\vy^{i}_{j}; \vx^{i}_{j}, \vtheta) \\
              = &\text{exp} \biggl( - \frac{1}{|\vy^{i}_{j}|} \sum_{k=1}^{|\vy^{i}_{j}|} \text{log}p_{\vtheta}(y_{j,k}|\vx^{i}_{j},\vy_{j,<k})\biggl).
    \end{aligned}
\end{equation}

However, we argue that perplexity is not a suitable metric for evaluating the difficulty of non-natural language texts, such as mathematical formulas and programming codes. Our preliminary study indicates that the perplexity scores given to mathematical texts by various language models are typically lower than those for natural language texts, despite the common belief that mathematical problems pose significant challenges for LLMs \citep{DBLP:journals/corr/abs-2309-05653,DBLP:journals/corr/abs-2309-12284}. Our preliminary study is presented in \autoref{sec:ppl}. Therefore, given a random mini-batch $B^{i} = \{ (\vx^{i}_{j}, \vy^{i}_{j}) \}_{j=1}^{L}$ from $\dtrain^{i}$, the reward $\mathcal{R}_{\diff}(i)$ for $\dtrain^{i}$ is:
\begin{align}
    \mathcal{R}_{\diff}(i) = \frac{1}{L} \sum^{L}_{j=1} \frac{\text{PPL}(\vy^{i}_{j}; \vx^{i}_{j}, \vtheta)}{\text{PPL}(\vy^{i}_{j}; \vx^{i}_{j}, \vtheta_{0})},
\end{align}
where $\vtheta_{0}$ is the original LLM backbone and $\vtheta$ is the fine-tuned LLM. 
The term $\mathcal{R}_{\diff}(i)$ represents the relative decrease in perplexity for $\dtrain^{i}$ after fine-tuning. A high value of $\mathcal{R}_{\diff}(i)$ suggests that $\dtrain^{i}$ is difficult to learn and requires more training efforts, while a lower value indicates the opposite.

\subsubsection{Learning Trajectory}
\label{sec:trajectory}

We design the rewards $\mathcal{R}_{\cossim}(i)$ and $\mathcal{R}_{\diff}(i)$ based on the transferability and difficulty of the training dataset $\dtrain^{i}$, as discussed in \autoref{sec:transfer} and \autoref{sec:difficult}. However, both rewards ignore the learning trajectory of the fine-tuning process. Therefore, we introduce the exponential moving average (\ema) when estimating the rewards. This approach can both better estimate the reward and stabilize the data usage optimization process. Specifically, we define the \ema as follows:
\begin{align}
    \label{eq:ema}
    \mathcal{R}_{\{\cdot\}}(i) = \beta \mathcal{R}^{'}_{\{\cdot\}}(i) + (1-\beta)\mathcal{R}^{''}_{\{\cdot\}}(i),
\end{align}
where $\beta$ is the smoothing factor, $\mathcal{R}^{'}_{\{\cdot\}}(i)$ indicates the original reward for the current update, $\mathcal{R}^{''}_{\{\cdot\}}(i)$ represents the reward for the previous update, and $\mathcal{R}_{\{\cdot\}}(i)$ is the smoothed reward for the current update. Note that both $\mathcal{R}_{\cossim}(i)$ and $\mathcal{R}_{\diff}(i)$ can be applied in \autoref{eq:ema} and we set $\beta=0.9$.

\section{Experiments}
\label{sec:experiments}
We present our experimental setup (\autoref{sec:setup}) and main results (\autoref{sec:main_results}) in this section.

\subsection{Experimental Setup}
\label{sec:setup}

\begin{table}[t]
\centering
\small
\setlength{\tabcolsep}{3pt}
\begin{tabular}{lccccc}
\toprule
                      & \#exam.       & \#words         & Inst.L      & Resp.L      & Turns \\ \midrule
\texttt{Mathematics} & \phantom{0}26.2K & \phantom{0}3.4M & \phantom{0}47.6 & \phantom{0}84.0 & 1.0   \\
\texttt{Medicine}  & \phantom{00}5.2K & \phantom{0}1.0M & \phantom{0}36.5 & 147.4           & 1.0   \\
\texttt{General}     & \phantom{00}9.3K & \phantom{0}9.3M & \phantom{0}54.3 & 202.4           & 3.6   \\
\texttt{NLP}           & \phantom{0}62.6K & \phantom{0}8.6M & 127.9           & \phantom{00}9.7 & 1.0   \\ \midrule
Total                 & 103.4K           & 22.3M           & \phantom{0}88.4 & \phantom{0}84.3 & 1.2  \\ \bottomrule
\end{tabular}
\caption{
Dataset statistics of the training datasets in this work. 
Inst.L, Resp.L, and Turns indicate the average of instruction length (in words), response length (in words), and number of conversation turns.
}
\label{tab:data_stat}
\end{table}
\paragraph{Datasets}
In this work, we collect four distinct supervised fine-tuning (SFT) datasets:
\begin{itemize}
    \item \textbf{\dataset{Mathematics}}: \citet{DBLP:journals/corr/abs-2309-05653} introduce \dataset{MathInstruct}, a comprehensive collection of mathematical SFT datasets.\footnote{\url{https://huggingface.co/datasets/TIGER-Lab/MathInstruct}}
    \item \textbf{\dataset{Medicine}}: \citet{DBLP:journals/corr/abs-2310-14558} introduce a medical SFT dataset \dataset{MedInstruct}.\footnote{\url{https://huggingface.co/datasets/casey-martin/MedInstruct}}
    \item \textbf{\dataset{General}}: The \dataset{ShareGPT} dataset serves as our general SFT dataset, contributed by the general public and characterized by a high degree of diversity and quality.\footnote{\url{https://huggingface.co/datasets/anon8231489123/ShareGPT_Vicuna_unfiltered}}
    \item \textbf{\dataset{NLP}}: \citet{DBLP:conf/iclr/SanhWRBSACSRDBX22} open-source \dataset{P3}, which is a collection of prompted English datasets covering a diverse set of NLP tasks.\footnote{\url{https://huggingface.co/datasets/bigscience/P3}} Note that the \dataset{P3} collection consists of 660 subsets, totaling 122M examples. To ensure task diversity, we initially randomly select 1K examples from each subset.
\end{itemize}
Due to the constraint on compute, we sample 10\% of examples from each dataset and present the dataset statistics in \autoref{tab:data_stat}.

% \usepackage{multirow}
\begin{table*}[t]
\centering
\small
\setlength{\tabcolsep}{10pt}
\begin{tabular}{lcccccccc}
\toprule
                       & \multirow{2}{*}{$\avgboth$} & \multicolumn{4}{c}{\textbf{\mmlu}}                                         & \multicolumn{3}{c}{\textbf{\mtbench}}                  \\ \cmidrule(rl){3-6} \cmidrule(rl){7-9}
                       &                             & $\avgmmlu$     & Math           & Med.            & Others         & $\avgmtbench$ & Turn 1        & Turn 2        \\ \midrule
\multicolumn{9}{c}{\cellcolor{gray!30}\qwen}  \\
\prop ($\tau=1$)       & 32.82                       & 30.95          & \textbf{23.40} & \underline{31.30}    & 31.76          & 3.47          & 4.18          & 2.76          \\
\temp ($\tau=10$)      & 34.17                       & 32.09          & \underline{22.88}    & 30.88          & 33.41          & 3.63          & 4.11          & 3.14          \\
\uni ($\tau=\infty$)   & 33.81                       & 31.52          & 21.45          & 29.97          & 33.02          & 3.61          & 4.21    & 3.01          \\ \hdashline
\multidds       & 33.67                       & 31.02          & 22.84    & 30.28          & 33.01          & 3.63          & 4.13          & 3.14          \\
\multiuat   & 34.15                       & 31.24          & 21.41          & 29.91          & 33.01          & 3.71          & \underline{4.26}    & 3.15          \\ \hdashline
\ours + \cossim        & 34.30                       & 31.95          & 21.90          & 31.18          & 33.28          & 3.67          & 3.96          & \underline{3.38}    \\
\ours + \cossim + \ema & \textbf{35.13}              & \textbf{32.45} & 22.27          & \textbf{31.56} & \textbf{33.82} & \textbf{3.78} & \textbf{4.44} & 3.13          \\
\ours + \diff          & 34.24                       & 31.49          & 20.71          & 29.94          & 33.07          & 3.70          & \underline{4.21}    & 3.19          \\
\ours + \diff + \ema   & \underline{34.83}                 & \underline{32.11}    & 21.84          & 31.01          & \underline{33.53}    & \underline{3.76}    & 4.11          & \textbf{3.40} \\ \midrule
\multicolumn{9}{c}{\cellcolor{gray!30}\gemma}  \\
\prop ($\tau=1$)       & 42.90                       & 33.61          & 20.03          & 33.33          & 35.25          & 5.22          & 5.63          & 4.81          \\
\temp ($\tau=10$)      & 41.86                       & \textbf{36.16} & \textbf{21.03} & \textbf{37.92} & \underline{37.55}    & 4.76          & 5.49          & 4.03          \\
\uni ($\tau=\infty$)   & 43.95                       & \underline{35.95}    & 20.82          & \underline{35.97}    & \textbf{37.71} & 5.19          & 5.60          & 4.79          \\ \hdashline
\multidds       & 41.11                       & 34.59          & 20.33    & 33.17          & 34.55          & 4.76          & 5.48          & 4.05          \\
\multiuat   & 43.19                       & 33.95          & 20.85          & 34.64          & 33.71          & 5.24          & 5.61    & 4.88          \\ \hdashline
\ours + \cossim        & 43.84                       & 32.44          & 20.61          & 32.35          & 33.83          & 5.53          & \textbf{6.16} & 4.89          \\
\ours + \cossim + \ema & 44.49                       & 33.86          & 20.57          & 33.32          & 35.52          & 5.51          & 6.01          & \underline{5.01}    \\
\ours + \diff          & \underline{44.93}                 & 34.32          & 20.29          & 34.06          & 36.00          & \underline{5.55}    & 6.04          & \textbf{5.06} \\
\ours + \diff + \ema   & \textbf{45.10}              & 34.61          & \underline{20.83}    & 34.64          & 36.20          & \textbf{5.56} & \underline{6.13}    & 4.99          \\ \midrule
\multicolumn{9}{c}{\cellcolor{gray!30}\llama}  \\
\prop ($\tau=1$)       & 60.97                       & 56.78          & 26.61          & 62.03          & 59.19          & 6.52          & 6.96          & 6.08          \\
\temp ($\tau=10$)      & 61.40                       & 56.17          & 28.36          & 59.64          & 58.68          & 6.66          & 7.04          & 6.29          \\
\uni ($\tau=\infty$)   & 60.99                       & 55.72          & 27.65          & 60.77          & 57.93          & 6.63          & 7.04          & 6.21          \\ \hdashline
\multidds       & 61.77	& 56.65	& 28.98	& 59.99	& 58.88	& 6.69	& 7.14	& 6.24          \\
\multiuat   & 61.18	& 55.66	& 28.65	& 60.48	& 57.32	& 6.67	& 7.05	& 6.29         \\ \hdashline
\ours + \cossim        & 62.49                       & 56.95          & 28.91          & 59.91          & 59.59          & 6.80          & 7.11          & 6.50          \\
\ours + \cossim + \ema & \textbf{63.85}              & \underline{58.08}    & 27.60          & 61.54          & \textbf{60.90} & \textbf{6.96} & \underline{7.28}    & \textbf{6.65} \\
\ours + \diff          & 63.00                       & 57.93          & \underline{31.08}    & \textbf{62.65} & 60.07          & 6.81          & 6.98          & \underline{6.64}    \\
\ours + \diff + \ema   & \underline{63.26}                 & \textbf{58.34} & \textbf{32.81} & \underline{62.21}    & \underline{60.49}    & \underline{6.82}    & \textbf{7.34} & 6.30         \\ \bottomrule
\end{tabular}

\caption{
Main results given by \qwen, \gemma, and \llama on \mmlu and \mtbench. The best and second-best results are highlighted in \textbf{bold} and \underline{underline}. Note that $\avgmtbench$ is upscaled by $10 \times$ to a range from 1 to 100 used for computing $\avgboth$.
}
\label{tab:main}
\end{table*}

\paragraph{Model Backbones}
We apply \ours to three diverse model backbones, including \qwen \citep{DBLP:journals/corr/abs-2309-16609}, \gemma \citep{DBLP:journals/corr/abs-2403-08295}, and \llama \citep{llama3modelcard}.

\paragraph{Optimization}
We fine-tune all the parameters of large language models (LLMs) using the AdamW optimizer \citep{DBLP:journals/corr/KingmaB14,DBLP:conf/iclr/LoshchilovH19} with a learning rate of $1\times10^{-5}$ and a batch size of 64. We fine-tune all the models for 3 epochs, or the equivalent number of steps. During the training process, we apply the linear learning rate schedule, which includes a warm-up phase comprising 10\% of the total training steps. For \ours, $\vscorer$ is updated for every 100 steps with the learning rate of $1 \times 10^{-4}$ and the batch size of 64. $\vscorer$ is initialized by $\tau = \infty$ in \autoref{eq:temp}. 

\paragraph{Baselines}
We compare \ours with several heuristic and dynamic baselines as follows:
\begin{itemize}
    \item \textbf{Heuristic}: Based on Equation~\ref{eq:temp}, we consider proportional sampling (\prop, $\tau = 1$), temperature sampling (\temp, $\tau = 10$), and uniform sampling (\uni, $\tau = \infty$).
    \item \textbf{Dynamic}: \multidds \citep{wang-etal-2020-balancing} and \multiuat \citep{wu-etal-2021-uncertainty} dynamically balance the dataset distribution using the gradient cosine similarity and model uncertainty on the validation sets as rewards, respectively. We sample 1K examples from each dataset as the validation sets for both approaches.
\end{itemize}
We do not include maximum cap and other instance selection methods as baselines because they do not fully utilize all available data.

\paragraph{Evaluation}
In this work, we evaluate two widely-used benchmarks that are highly correlated with human judgments:
\begin{itemize}
    \item \textbf{\mmlu}: \citet{DBLP:conf/iclr/HendrycksBBZMSS21} propose the \mmlu benchmark, covering 57 subjects across STEM, humanities, social sciences, and more. We categorize the subjects into three groups: mathematics, medicine, and others, and \textbf{conduct zero-shot evaluations}. We report the average accuracy for each group and the accuracy across all subjects, denoted as $\avgmmlu$. Detailed subject categorization is in \autoref{sec:mmlu_cate}.
    \item \textbf{\mtbench}: \citet{DBLP:journals/corr/abs-2306-05685} propose the \mtbench, a multi-turn conversational benchmark designed to measure large language models' capabilities. This benchmark covers eight key skills, including coding, writing, roleplay, and more. LLM responses are scored by \model{gpt-4} on a scale from 1 to 10. The overall score across all eight skills is denoted as $\avgmtbench$.
\end{itemize}
The overall performance is reported as the average score of both $\avgmmlu$ and $\avgmtbench$, denoted as $\avgboth$. Note that when computing $\avgboth$, \mtbench scores are upscaled by $10 \times$ to range from 1 to 100, maintaining consistency with \mmlu.

\subsection{Main Results}
\label{sec:main_results}
We present the main results in \autoref{tab:main}.

\paragraph{An optimal temperature $\tau$ boosts performance, but no universally optimal $\tau$ exists.}
When comparing the heuristic baselines, we observe that there is no universally optimal $\tau$ that consistently works well for all model backbones. As shown in \autoref{tab:main}, \temp ($\tau=10$) performs best for \qwen and \llama, but is least effective for \gemma. This variability confirms the motivation behind this work.

\paragraph{\ours outperforms heuristic baselines, with larger models showing greater improvements.}
Our method consistently outperforms heuristic baselines across all three model backbones in terms of $\avgboth$. Notably, larger models show greater improvements with our approach. As shown in \autoref{tab:main}, the best variant of our method surpasses the best heuristic baseline by +0.96, +1.15, and +2.45 in $\avgboth$ for \qwen, \gemma, and \llama, respectively. This is particularly significant in the era of \textbf{LARGE} language models. 

\paragraph{Different rewards work better for different models, and \ema always helps.}
As shown in \autoref{tab:main}, \ours with \cossim outperforms \ours with \diff for \qwen and \llama, while \diff-based \ours yields better results for \gemma. Additionally, \ema consistently enhances overall performance in terms of $\avgboth$, supporting our rationales concerning learning trajectory in \autoref{sec:trajectory}.

\begin{figure}[t]
    \centering
    
\begin{subfigure}[t]{0.23\textwidth}
\centering
    \begin{tikzpicture}
    \begin{axis}[
        xlabel={\small Trainning Steps},
        ylabel={\small Samp. Prob.},
        ymax=0.38,ymin=0.17,
        grid=both,
        grid style={line width=.1pt, draw=gray!30},
        height=4cm,
        width=\textwidth,
        tick label style={font=\scriptsize},
        legend style={,
            at={(0.5,1.4)},
            anchor=north west,
            legend columns=2,
            font=\scriptsize
        },
        legend cell align={left}, 
        y tick label style={
            /pgf/number format/.cd,
            fixed,
            fixed zerofill,
            precision=2,
            /tikz/.cd
        },
        no markers,
    ]
    \addplot[red] table [x=step, y=math, col sep=comma, mark=none] {figures/Meta-Llama-3-8B-dyna-t-inf-r-cossim-u-100-data-small-general_math_medical_p3-ema-yes-0.9.csv};
    \addplot[blue] table [x=step, y=medical, col sep=comma, mark=none] {figures/Meta-Llama-3-8B-dyna-t-inf-r-cossim-u-100-data-small-general_math_medical_p3-ema-yes-0.9.csv};
    \addplot[green] table [x=step, y=general, col sep=comma, mark=none] {figures/Meta-Llama-3-8B-dyna-t-inf-r-cossim-u-100-data-small-general_math_medical_p3-ema-yes-0.9.csv};
    \addplot[orange] table [x=step, y=p3, col sep=comma, mark=none] {figures/Meta-Llama-3-8B-dyna-t-inf-r-cossim-u-100-data-small-general_math_medical_p3-ema-yes-0.9.csv};
    \legend{\texttt{Mathematics}, \texttt{Medicine}, \texttt{General}, \texttt{NLP}}
    \end{axis}
    \end{tikzpicture}
    \caption{\ours + \cossim + \ema}

\end{subfigure}
~
\begin{subfigure}[t]{0.23\textwidth}
\centering
    \begin{tikzpicture}
    \begin{axis}[
        xlabel={\small Trainning Steps},
        ylabel={\small Samp. Prob.},
        ymax=0.38,ymin=0.17,
        grid=both,
        grid style={line width=.1pt, draw=gray!30},
        height=4cm,
        width=\textwidth,
        tick label style={font=\scriptsize},
        legend style={,
            at={(-0.15,-0.35)},
            anchor=north,
            legend columns=1,
            font=\small
        },
        y tick label style={
            /pgf/number format/.cd,
            fixed,
            fixed zerofill,
            precision=2,
            /tikz/.cd
        },
        no markers,
    ]
    \addplot[red] table [x=step, y=math, col sep=comma, mark=none] {figures/Meta-Llama-3-8B-dyna-t-inf-r-learn-u-100-data-small-general_math_medical_p3-ema-yes-0.9.csv};
    \addplot[blue] table [x=step, y=medical, col sep=comma, mark=none] {figures/Meta-Llama-3-8B-dyna-t-inf-r-learn-u-100-data-small-general_math_medical_p3-ema-yes-0.9.csv};
    \addplot[green] table [x=step, y=general, col sep=comma, mark=none] {figures/Meta-Llama-3-8B-dyna-t-inf-r-learn-u-100-data-small-general_math_medical_p3-ema-yes-0.9.csv};
    \addplot[orange] table [x=step, y=p3, col sep=comma, mark=none] {figures/Meta-Llama-3-8B-dyna-t-inf-r-learn-u-100-data-small-general_math_medical_p3-ema-yes-0.9.csv};
    % \legend{\texttt{Math}, \texttt{Med.}, \texttt{Gen.}, \texttt{NLP}}
    \end{axis}
    \end{tikzpicture}
    \caption{\ours + \diff + \ema}

\end{subfigure}
\caption{
Learned dataset distribution given by \llama with different variations of \ours.
The $x$-axis indicates the training steps, and the $y$-axis indicates the sampling probabilities of datasets. 
}
\label{fig:learned_dist}
\end{figure}
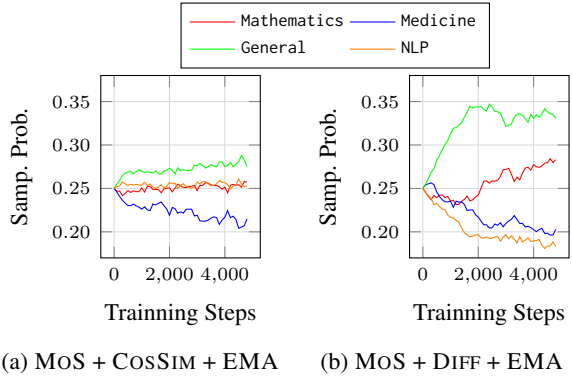

\section{Analysis}
\label{sec:analysis}

In this section, we conduct an in-depth analysis of \ours using \llama. Our analysis encompasses the learned dataset distribution of \ours, the training convergence speed of \ours, the impact of sampling priors on \ours, and its compatibility with the instance selection methods.

\paragraph{\ours with different rewards learns different dataset distributions.}
We visualize the dataset distribution learned by \ours using \llama as shown in \autoref{fig:learned_dist}. Starting with equal sampling probabilities, both \cossim and \diff variations of \ours increase the probability for \dataset{General} and decrease it for \dataset{Medicine}. However, \cossim maintains the probabilities for \dataset{Mathematics} and \dataset{NLP}, whereas \diff upsamples \dataset{Mathematics} but downsamples \dataset{NLP}.

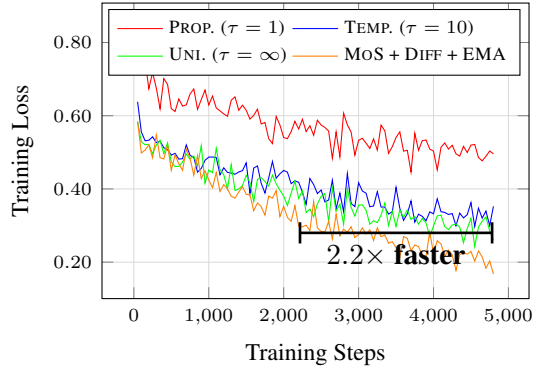
\begin{figure}[t]
\centering
\begin{tikzpicture}
\begin{axis}[
    xlabel={\small Training Steps},
    ylabel={\small Training Loss},
    grid=both,
    grid style={line width=.1pt, draw=gray!30},
    height=5.5cm,
    width=0.45\textwidth,
    tick label style={font=\scriptsize},
    legend style={,
        anchor=north east,
        legend columns=2,
        font=\scriptsize,
    },
    legend cell align={left}, 
    y tick label style={
        /pgf/number format/.cd,
        fixed,
        fixed zerofill,
        precision=2,
        /tikz/.cd
    },
    no markers,
]
\addplot[red] table [x=step, y=prop, col sep=comma, mark=none] {figures/train_loss.csv};
\addplot[blue] table [x=step, y=temp, col sep=comma, mark=none] {figures/train_loss.csv};
\addplot[green] table [x=step, y=uni, col sep=comma, mark=none] {figures/train_loss.csv};
\addplot[orange] table [x=step, y=learn, col sep=comma, mark=none] {figures/train_loss.csv};
\legend{\prop ($\tau=1$), \temp ($\tau=10$), \uni ($\tau=\infty$), \ours + \diff + \ema}

\draw[|-|, line width=1pt] (2200,0.28) -- (4800,0.28) ;
\node[] at (3500,0.22) {\bf $2.2\times$ faster};

\end{axis}
\end{tikzpicture}
\caption{Training loss curves of heuristic baselines and \ours + \diff + \ema.}
\label{fig:train_loss}
\end{figure}
\paragraph{\ours speeds up the convergence.}
We illustrate the training dynamics of heuristic baselines and \ours + \diff + \ema in \autoref{fig:train_loss}. When compared to \temp ($\tau = 10$), which is the best-performing heuristic baseline, \ours + \diff + \ema demonstrates notable improvements. Specifically, it converges approximately $2.2 \times$ faster, as shown in \autoref{fig:train_loss}, and achieves a +1.86 improvement in terms of $\avgboth$, as detailed in \autoref{tab:main}.

\begin{table}[t]
\centering
\small
\setlength{\tabcolsep}{10pt}
\begin{tabular}{lccc}
\toprule
                     & $\avgboth$ & $\avgmmlu$ & $\avgmtbench$ \\ \midrule
\temp ($\tau=10$)     & 61.40      & 56.17      & 6.66          \\ \midrule
\multicolumn{4}{l}{\cellcolor{gray!30}\ours + \cossim + \ema}  \\
\quad + $\tau=1$      & 62.49      & 56.95      & 6.80          \\
\quad + $\tau=10$     & \textbf{63.94}      & 57.98      & \textbf{6.99}          \\
\quad + $\tau=\infty$ & 63.85      & \textbf{58.08}     & 6.96          \\ \midrule
\multicolumn{4}{l}{\cellcolor{gray!30}\ours + \diff + \ema}    \\
\quad + $\tau=1$      & 62.29      & 56.51      & 6.81          \\
\quad + $\tau=10$     & \textbf{63.66}      & 58.22      & \textbf{6.91}          \\
\quad + $\tau=\infty$ & 63.26      & \textbf{58.34}      & 6.82          \\ \bottomrule
\end{tabular}
\caption{
Results of \ours with different sampling priors $\tau$.
The best results are highlighted in \textbf{bold}.
}
\label{tab:prior}
\end{table}
\paragraph{\ours demonstrates robustness to changes in sampling priors.}
As indicated in \autoref{line:line1} in \autoref{alg:mos}, we initialize our sampling probability distribution with $\tau=\infty$. Consequently, we investigate the effects of various sampling priors on \ours. As shown in \autoref{tab:prior}, \ours with different sampling priors consistently outperforms the best heuristic baseline \temp ($\tau=10$). Moreover, selecting the appropriate sampling prior for \ours can further enhance performance. These results underscore the robustness and effectiveness of our approach.

\begin{table}[t]
\centering
\small
\setlength{\tabcolsep}{7pt}
\begin{tabular}{lccc}
\toprule
                       & $\avgboth$ & $\avgmmlu$ & $\avgmtbench$ \\ \midrule
\multicolumn{4}{l}{\cellcolor{gray!30}\model{Random}}                               \\
\prop ($\tau=1$)       & 59.78      & 54.39      & 6.52          \\
\temp ($\tau=10$)      & 60.62      & 54.89      & 6.63          \\
\uni ($\tau=\infty$)   & 59.21      & 54.21      & 6.42          \\ \midrule
\multicolumn{4}{l}{\cellcolor{gray!30}\model{IFD}}                                  \\
\prop ($\tau=1$)       & 60.43      & 55.03      & 6.58          \\
\temp ($\tau=10$)      & 61.02      & 55.13      & 6.69          \\
\uni ($\tau=\infty$)   & 60.62      & 54.92      & 6.63          \\ \hdashline
\ours + \cossim + \ema & 62.01      & \textbf{56.81}      & 6.72          \\
\ours + \diff + \ema   & \textbf{62.05}      & 56.43      & \textbf{6.77} \\ \bottomrule
\end{tabular}
\caption{
Compatibility between \ours and \model{IFD}. \model{Random} and \model{IFD} indicate the 10\% data from each dataset selected by random sampling and \model{IFD} selection, respectively.
The best results are highlighted in \textbf{bold}.
}
\label{tab:ifd}
\end{table}
\paragraph{\ours is compatible with the instance selection method.} Following \citet{DBLP:journals/corr/abs-2308-12032}, we leverage \qwen to calculate the Instruction-Following Difficulty (\model{IFD}) scores for each training instance and select the top 10\% of training instances with the highest scores from each dataset. As shown in \autoref{tab:ifd}, combining \ours with \model{IFD} further improves the model performance, indicating the successful combination of \ours and \model{IFD}.

\section{Fine-tuning from Generalist to Specialist}
\label{sec:ourspec}
\begin{table}[t]
\centering
\small
\setlength{\tabcolsep}{3pt}
\begin{tabular}{lcccc}
\toprule
                                       & \multirow{2}{*}{$\avgall$} & \dataset{GSM8K} & \dataset{MATH}  & \dataset{M-math} \\
                                       &                            & 5-shot          & 5-shot          & 0-shot           \\ \midrule
\multicolumn{5}{l}{\cellcolor{gray!30}\textit{Generalist}}                                                                                    \\
\temp ($\tau=10$)                      & 29.17                      & 49.62           & \phantom{0}9.54 & 28.36            \\
\ours + \cossim + \ema                 & 29.26                      & 50.40           & \phantom{0}9.78 & 27.60            \\
\ours + \diff + \ema                   & 30.88                      & 49.58           & 10.26           & \textbf{32.81}   \\ \midrule
\multicolumn{5}{l}{\cellcolor{gray!30}\textit{Math Specialist}}                                                                               \\
\mathllama                             & 27.04                      & 41.02           & \phantom{0}9.76 & 30.34            \\
\ourspec + \textsc{C.S.} + \textsc{E.} & 30.63                      & 51.10           & 10.64           & 30.16            \\
\ourspec + \textsc{D.} + \textsc{E.}   & \textbf{31.91}             & \textbf{52.10}  & \textbf{11.40}  & 32.24            \\ \bottomrule
\end{tabular}
\caption{
Results on \dataset{GSM8K}, \dataset{MATH}, and \dataset{M-math} given by generalists and math specialists.
$\avgall$ indicates the average performance over all three benchmarks.
\ourspec + \textsc{C.S.} + \textsc{E.} and \ourspec + \textsc{D.} + \textsc{E.} indicates \ourspec + \cossim + \ema and \ourspec + \diff + \ema, respectively.
The best results are highlighted in \textbf{bold}.
}
\label{tab:math}
\end{table}

Large, general-purpose models offer broad capabilities but can be costly to deploy in real-world applications. Many scenarios require only a narrow set of functionalities, making smaller, specialized models more effective for specific tasks than larger, general-purpose ones \citep{DBLP:journals/corr/abs-2306-08568, DBLP:journals/corr/abs-2310-10631, DBLP:journals/corr/abs-2401-06468}. \oursfull (\ours) is a framework designed to optimize data usage for various fine-tuning purposes, including task-specific fine-tuning. This section explores the application of \ours in this context.

We aim to fine-tune \llama to specialize in mathematics using datasets from \autoref{sec:setup}, referring to this modified version as \ourspec. For \ourspec with \cossim, we compute the cosine similarity between \dataset{Mathematics} and other datasets, including \dataset{Mathematics} itself. For \ourspec with \diff, we double the reward for the \dataset{Mathematics} dataset. For comparison, we fine-tune \llama directly on \dataset{Mathematics} dataset in \autoref{sec:setup}, denoted as \mathllama. Both \ourspec and \mathllama use the identical hyperparameters from \autoref{sec:setup}, except \mathllama is fine-tuned for 12 epochs for fairness.  We evaluate the models on math-related subjects in \mmlu (0-shot, denoted as \dataset{M-math}), \dataset{GSM8K} (5-shot) \citep{DBLP:journals/corr/abs-2110-14168}, and \dataset{MATH} (5-shot) \citep{DBLP:conf/nips/HendrycksBKABTS21}. 

\paragraph{SFT datasets from other sources are beneficial for the specific target task.}
As shown in \autoref{tab:math}, the \mathllama model trained solely on the \dataset{Mathematics} subset performs the worst among all models. This indicates that incorporating additional SFT datasets is advantageous. The performance gap is particularly evident on the \dataset{GSM8K} dataset, which requires step-by-step reasoning. We believe this discrepancy arises from the \dataset{Mathematics} subset's incompleteness, while other SFT datasets can address these shortcomings.

\begin{figure}[t]
\centering

\begin{subfigure}[t]{0.23\textwidth}
\centering
    \begin{tikzpicture}
    \begin{axis}[
        xlabel={\small Trainning Steps},
        ylabel={\small Samp. Prob.},
        ymax=0.45,ymin=0.10,
        grid=both,
        grid style={line width=.1pt, draw=gray!30},
        height=4cm,
        width=\textwidth,
        tick label style={font=\scriptsize},
        legend style={,
            at={(0.5,1.4)},
            anchor=north west,
            legend columns=2,
            font=\scriptsize
        },
        legend cell align={left}, 
        y tick label style={
            /pgf/number format/.cd,
            fixed,
            fixed zerofill,
            precision=2,
            /tikz/.cd
        },
        no markers,
    ]
    \addplot[red] table [x=step, y=math, col sep=comma, mark=none] {figures/Meta-Llama-3-8B-dyna-t-inf-r-cossim-u-100-data-small-general_math_medical_p3-ema-yes-0.9-included-spec-math.csv};
    \addplot[blue] table [x=step, y=medical, col sep=comma, mark=none] {figures/Meta-Llama-3-8B-dyna-t-inf-r-cossim-u-100-data-small-general_math_medical_p3-ema-yes-0.9-included-spec-math.csv};
    \addplot[green] table [x=step, y=general, col sep=comma, mark=none] {figures/Meta-Llama-3-8B-dyna-t-inf-r-cossim-u-100-data-small-general_math_medical_p3-ema-yes-0.9-included-spec-math.csv};
    \addplot[orange] table [x=step, y=p3, col sep=comma, mark=none] {figures/Meta-Llama-3-8B-dyna-t-inf-r-cossim-u-100-data-small-general_math_medical_p3-ema-yes-0.9-included-spec-math.csv};
    \legend{\textbf{\texttt{Mathematics}}, \texttt{Medicine}, \texttt{General}, \texttt{NLP}}
    \end{axis}
    \end{tikzpicture}
    % \caption{}
    \label{fig:mospec_cossim}
\end{subfigure}
~
\begin{subfigure}[t]{0.23\textwidth}
\centering
    \begin{tikzpicture}
    \begin{axis}[
        xlabel={\small Trainning Steps},
        ylabel={\small Samp. Prob.},
        ymax=0.45,ymin=0.10,
        grid=both,
        grid style={line width=.1pt, draw=gray!30},
        height=4cm,
        width=\textwidth,
        tick label style={font=\scriptsize},
        legend style={,
            at={(-0.4,1.3)},
            anchor=north west,
            legend columns=2,
            font=\small
        },
        y tick label style={
            /pgf/number format/.cd,
            fixed,
            fixed zerofill,
            precision=2,
            /tikz/.cd
        },
        no markers,
    ]
    \addplot[red] table [x=step, y=math, col sep=comma, mark=none] {figures/Meta-Llama-3-8B-dyna-t-inf-r-learn-u-100-data-small-general_math_medical_p3-ema-yes-0.9-included-spec-math.csv};
    \addplot[blue] table [x=step, y=medical, col sep=comma, mark=none] {figures/Meta-Llama-3-8B-dyna-t-inf-r-learn-u-100-data-small-general_math_medical_p3-ema-yes-0.9-included-spec-math.csv};
    \addplot[green] table [x=step, y=general, col sep=comma, mark=none] {figures/Meta-Llama-3-8B-dyna-t-inf-r-learn-u-100-data-small-general_math_medical_p3-ema-yes-0.9-included-spec-math.csv};
    \addplot[orange] table [x=step, y=p3, col sep=comma, mark=none] {figures/Meta-Llama-3-8B-dyna-t-inf-r-learn-u-100-data-small-general_math_medical_p3-ema-yes-0.9-included-spec-math.csv};
    % \legend{\texttt{Mathematics}, \texttt{Medicine}, \texttt{General}, \texttt{NLP}}
    \end{axis}
    \end{tikzpicture}
    \label{fig:mospec_diff}

\end{subfigure}
\caption{
Learned dataset distribution given by \llama with \ourspec + \cossim + \ema (left) and \ourspec + \diff + \ema (right).
}
\label{fig:mospec_learned_dist}
\end{figure}
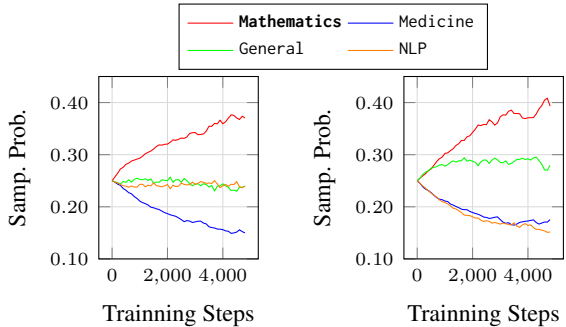
\paragraph{\ourspec can harness diverse datasets to enhance task-specific performance.}
When comparing \ours and \ourspec with the same reward type, \ourspec consistently outperforms \ours on mathematical benchmarks in \autoref{tab:math}. By assigning a higher reward, \ourspec effectively learns the optimal dataset distribution for learning the mathematical capabilities. As shown in \autoref{fig:mospec_learned_dist}, \cossim and \diff in \ourspec effectively guide the scorer network $\vscorer$ to increase the sampling probability of \dataset{Mathematics}.

We believe this property of \ours is particularly meaningful when the task-specific dataset is not sufficiently large or comprehensive.

\section{Related Work}

\paragraph{Data Engineering for LLMs}
The success of large language models (LLMs) heavily relies on their training datasets. Researchers gather or create extensive datasets \citep{DBLP:journals/jmlr/RaffelSRLNMZLL20,DBLP:journals/corr/abs-2101-00027,DBLP:conf/nips/PenedoMHCACPAL23,wang-etal-2023-self-instruct,DBLP:journals/corr/abs-2305-15011,DBLP:journals/corr/abs-2310-01377,wu-etal-2024-lamini,wang-etal-2024-retrieval}. Recent efforts focus on selecting data subsets to enhance training efficiency. \citet{DBLP:conf/nips/Xie0DDLLLL0Y23} estimate the quality of each subset in the pretraining dataset mixture using a small proxy model. Recent approaches filter out low-quality examples using perplexity \citep{DBLP:journals/corr/abs-2308-12032,DBLP:journals/corr/abs-2309-04564}.

\paragraph{Dataset Rebalancing}
The standard practice for dataset rebalancing in fine-tuning large language models (LLMs) involves capping the number of examples per dataset  \citep{DBLP:journals/jmlr/RaffelSRLNMZLL20, DBLP:journals/corr/abs-2210-11416, DBLP:journals/corr/abs-2212-12017, DBLP:conf/iclr/WeiBZGYLDDL22}. However, this approach does not fully utilize all available data. Previous multilingual research often rebalances datasets for multiple languages using a temperature term $\tau$ \citep{DBLP:journals/corr/abs-1907-05019, conneau-etal-2020-unsupervised}. Furthermore, \citet{DBLP:conf/icml/WangPMACN20} reweigh training examples based on their similarity with the validation set. Inspired by \citet{DBLP:conf/icml/WangPMACN20}, \citet{wang-etal-2020-balancing} and \citet{wu-etal-2021-uncertainty} propose rebalancing the dataset distribution for machine translation tasks.

\paragraph{Multi-Task Learning}
Our work is also related to multi-task learning \citep{DBLP:journals/corr/Ruder17a, DBLP:journals/corr/abs-2009-09796, zhang-etal-2023-survey}. Both transferability and difficulty are commonly used for reweighting the importance of tasks to achieve better overall performance and mitigate the conflicts between tasks \citep{DBLP:conf/cvpr/KendallGC18, DBLP:conf/icml/ChenBLR18, DBLP:conf/nips/YuK0LHF20, DBLP:conf/iclr/WangTF021}. We highlight that \textit{tasks} are the specific goals the model works towards, while \textit{skills} are the broader abilities that allow the model to perform a wide range of tasks.

\paragraph{Ours}
In this work, \oursfull (\ours) is inspired by \citet{DBLP:conf/icml/WangPMACN20} and related to \citet{wang-etal-2020-balancing} and \citet{wu-etal-2021-uncertainty}, but offers several key advancements. Unlike previous methods, \ours does not require knowledge of downstream applications, avoiding the risk of overfitting to validation sets. Additionally, \ours introduces novel rewards tailored for LLMs and considers the learning trajectory during fine-tuning, enhancing overall performance. Finally, \ours is highly adaptable for specific fine-tuning needs, setting it apart from prior works.

\section{Conclusion}

In this work, we address the critical challenge of optimizing data usage during the fine-tuning process of LLMs. We propose a general, model-agnostic reinforcement learning framework, \oursfull (\ours), that dynamically adjusts dataset usage to enhance model performance with three novel rewards. Through extensive experiments on three diverse model backbones and two widely-used benchmarks, we demonstrate that \ours significantly improves overall model performance. Additionally, we explore the application of \ours in task-specific fine-tuning, leading to the development of \ourspec. Our experiments show that models fine-tuned with \ourspec on various datasets outperform those trained solely on task-specific datasets. In summary, \ours provides a powerful and flexible solution to the challenges of dataset heterogeneity and imbalance in the fine-tuning of LLMs.

\section{Limitations}

\paragraph{Computational Overhead}
In this study, the scorer network $\vscorer$ and the large language model (LLM) $\vtheta$ are updated in an alternating fashion. Although the scorer network $\vscorer$ is a relatively simple two-layer MLP model, the overall training duration increases by approximately 20\%, compared with the heuristic baselines, when the LLM $\vtheta$ is updated for the same number of steps. 

\paragraph{Number of Datasets}
Our experiments are limited to four datasets due to computational resource constraints. The performance of our approach as the dataset count increases remains unexplored.

These limitations are acknowledged and we leave them to the future work.

\section*{Acknowledgment}
This work is partly supported by the ARC Future Fellowship FT190100039. This material is based on research partially supported by the DARPA Assured Neuro Symbolic Learning and Reasoning (ANSR) program under award number FA8750-23-2-1016.

\bibliography{custom}

\clearpage
\appendix

\begin{table*}[t]
\centering
\small
\setlength{\tabcolsep}{10pt}
\begin{tabular}{lccccccccc}
\toprule
            & \multicolumn{3}{c}{\qwen}                      & \multicolumn{3}{c}{\gemma}                     & \multicolumn{3}{c}{\llama}                     \\ \cmidrule(rl){2-4} \cmidrule(rl){5-7} \cmidrule(rl){8-10}
            & $\text{PPL}_{\vtheta_0}$ & $\text{PPL}_{\vtheta}$ & $\Delta$ & $\text{PPL}_{\vtheta_0}$ & $\text{PPL}_{\vtheta}$ & $\Delta$ & $\text{PPL}_{\vtheta_0}$ & $\text{PPL}_{\vtheta}$ & $\Delta$ \\ \midrule
\dataset{Mathematics} & \phantom{0}4.18   & 2.94            & 0.70     & \phantom{0}5.85   & 2.31            & 0.39     & \phantom{0}5.65   & 2.94            & 0.52     \\
\dataset{Medicine}    & \phantom{0}8.38   & 4.45            & 0.53     & \phantom{0}8.60   & 2.95            & 0.34     & \phantom{0}5.86   & 2.51            & 0.43     \\
\dataset{General}     & \phantom{0}6.05   & 4.01            & 0.66     & \phantom{0}9.66   & 3.51            & 0.36     & \phantom{0}4.25   & 2.51            & 0.59     \\
\dataset{NLP}         & 37.70             & 7.98            & 0.21     & 49.28             & 4.78            & 0.10     & 29.79             & 4.19            & 0.14     \\ \bottomrule
\end{tabular}
\caption{
Preliminary results on perplexity.
$\text{PPL}_{\vtheta_0}$ and $\text{PPL}_{\vtheta}$ are the average perplexity scores on each subset given by the original LLM backbone and the fine-tuned LLM with \prop ($\tau=1$), respectively.
$\Delta = \frac{\text{PPL}_{\vtheta}}{\text{PPL}_{\vtheta_0}}$ indicates the relative decrease in perplexity. 
A high value of $\Delta$ indicates the dataset is difficult to learn, while a lower value indicates the opposite.
}
\label{tab:prelim_ppl}
\end{table*}
\section{Preliminary Study on Perplexity}
\label{sec:ppl}

Perplexity measures how well a probability model predicts a sample, quantifying the model's uncertainty in making these predictions. It is designed for natural language texts because it relies on the probability distributions typical to human languages.  However, non-natural language texts, such as mathematical formulas or programming code, often involve symbols and structures whose relationships are governed by logical or mathematical rules rather than linguistic context. We hypothesize that using perplexity to measure the difficulty in these contexts does not capture the essential aspects of understanding or generating such texts.

To verify our hypothesis, we conduct a preliminary study on perplexity and present the results in \autoref{tab:prelim_ppl}. We observe that the perplexity of \dataset{Mathematics} is commonly lower than that of other datasets given by \qwen, \gemma, and \llama, regardless of whether the models are fine-tuned, while the perplexity of \dataset{NLP} is the highest among all the datasets. However, \dataset{Mathematics} is associated with a higher value of $\Delta$, while \dataset{NLP} achieves the lowest value of $\Delta$, suggesting that \dataset{Mathematics} is difficult to learn while \dataset{NLP} is easy to learn. If we utilize the perplexity as a measure of difficulty in \autoref{sec:difficult}, \ours incorrectly assigns a higher sampling probability to \dataset{NLP}.

\begin{table*}[t]
\centering
\small
\begin{tabular}{@{}lp{0.8\textwidth}@{}}
\toprule
            & Subjects                                                                                                                                                                                                                                                                                                                                                                                                                                                                                                                                                                                                                                                                                                                                                                                                                                                                    \\ \midrule
Mathematics & Abstract Algebra, College Mathematics, Elementary Mathematics, High School Mathematics, High School Statistics                                                                                                                                                                                                                                                                                                                                                                                                                                                                                                                                                                                                                                                                                                                                                              \\ \midrule
Medicine    & Anatomy, Clinical Knowledge, College Medicine, Human Aging, Human Sexuality, Medical Genetics, Nutrition, Professional Medicine, Virology                                                                                                                                                                                                                                                                                                                                                                                                                                                                                                                                                                                                                                                                                                                                   \\ \midrule
Others      & Astronomy, Business Ethics, College Biology, College Chemistry, College Computer Science, College Physics, Computer Security, Conceptual Physics, Econometrics, Electrical Engineering, Formal Logic, Global Facts, High School Biology, High School Chemistry, High School Computer Science, High School European History, High School Geography, High School Government And Politics, High School Macroeconomics, High School Microeconomics, High School Physics, High School Psychology, High School US History, High School World History, International Law, Jurisprudence, Logical Fallacies, Machine Learning, Management, Marketing, Miscellaneous, Moral Disputes, Moral Scenarios, Philosophy, Prehistory, Professional Accounting, Professional Law, Professional Psychology, Public Relations, Security Studies, Sociology, US Foreign Policy, World Religions \\ \bottomrule
\end{tabular}
\caption{\mmlu subject categorization.}
\label{tab:mmlu_cate}
\end{table*}
\section{MMLU Subject Categorization}
\label{sec:mmlu_cate}
We present the detailed subject categorization of \mmlu in \autoref{tab:mmlu_cate}.

\end{document}